\documentclass[11pt,a4paper]{article}
\usepackage[hyperref]{acl2019}
\usepackage{times}
\usepackage{latexsym}

\usepackage{url}

\usepackage{graphicx}
\usepackage{amsmath}
\usepackage{todonotes}
\usepackage{multirow, tabu}
\usepackage{comment}

\aclfinalcopy % Uncomment this line for the final submission
\def\aclpaperid{136} %  Enter the acl Paper ID here

\newcommand*{\affaddr}[1]{#1}
\newcommand*{\affmark}[1][*]{\textsuperscript{#1}}

\title{Tree-Structured Semantic Encoder with Knowledge Sharing \\ for Domain Adaptation in Natural Language Generation}
\author{
    Bo-Hsiang Tseng,\affmark[$\dagger$]
    Pawe\l\ Budzianowski,\affmark[$\dagger$]
    Yen-Chen Wu\affmark[$\dagger$]
    Milica~Ga{\v s}i{\' c}\affmark[$\ddagger$] \\
    \affaddr{\affmark[$\dagger$]University of Cambridge} \affaddr{\affmark[$\ddagger$] Heinrich Heine Univeristy D{\"u}sseldorf} \\
    \tt bht26@cam.ac.uk, gasic@uni-duesseldorf.de
}

\begin{document}
\maketitle
\begin{abstract}
Domain adaptation in natural language generation (NLG) remains challenging because of the high complexity of input semantics across domains and limited data of a target domain. This is particularly the case for dialogue systems, where we want to be able to seamlessly include new domains into the conversation. Therefore, it is crucial for generation models to share knowledge across domains for the effective adaptation from one domain to another. In this study, we exploit a tree-structured semantic encoder to capture the internal structure of complex semantic representations required for multi-domain dialogues in order to facilitate knowledge sharing across domains. In addition, a layer-wise attention mechanism between the tree encoder and the decoder is adopted to further improve the model's capability. The automatic evaluation results show that our model outperforms previous methods in terms of the BLEU score and the slot error rate, in particular when the adaptation data is limited. In subjective evaluation, human judges tend to prefer the sentences generated by our model, rating them more highly on informativeness and naturalness than other systems.
\end{abstract}

\section{Introduction}
\begin{figure}[tb]
  \includegraphics[width=\linewidth]{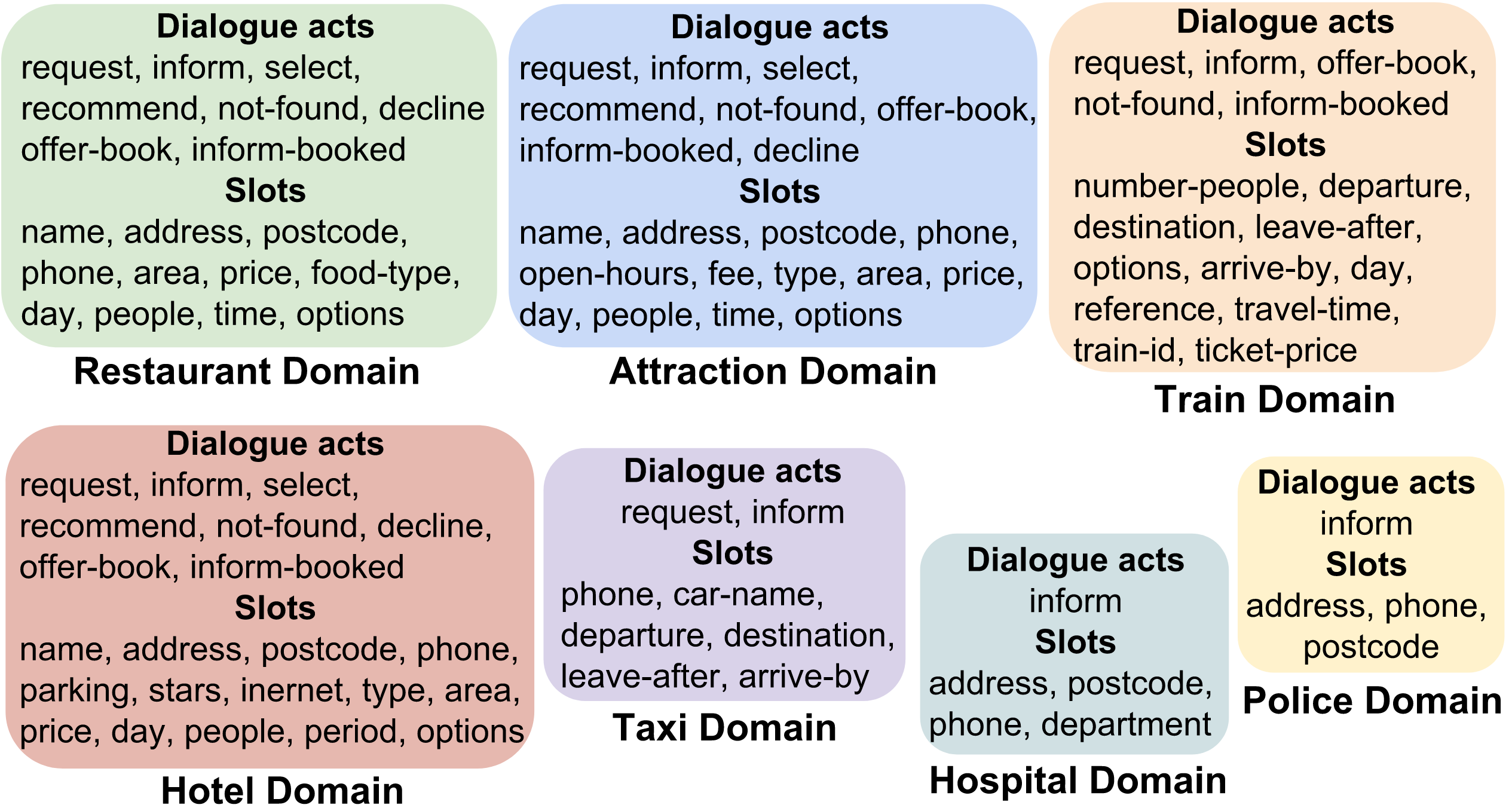}
  \caption{The ontology for multi-domain spoken dialogue systems.}
  \label{fig:ontology}
    \vspace{-0.5em}
\end{figure}

% Intro of SDS, adaptation studies in SDS
Building open-domain Spoken Dialogue Systems (SDS) remains challenging. This is partially because of the difficulty of collecting sufficient data for all domains and the high complexity of natural language. Typical SDSs are designed based on a pre-defined ontology (Figure \ref{fig:ontology}) which might cover knowledge spanning over multiple domains and topics \cite{young2013pomdp}. %In order to achieve acceptable performance in domains with limited data, a right structure is essential to share effectively the knowledge across domains. % is important for domain adaptation.
%a universal model that is capable of handling information from various domains and further sharing knowledge is required. Statistical approaches to SDS provide a good framework to build a multi-domain model. Instead of requiring substantial amounts of handcrafted rules, the flexibility of the data-driven approach allows models to adapt to new domains effectively.

\begin{figure}[tb]
  \includegraphics[width=\linewidth]{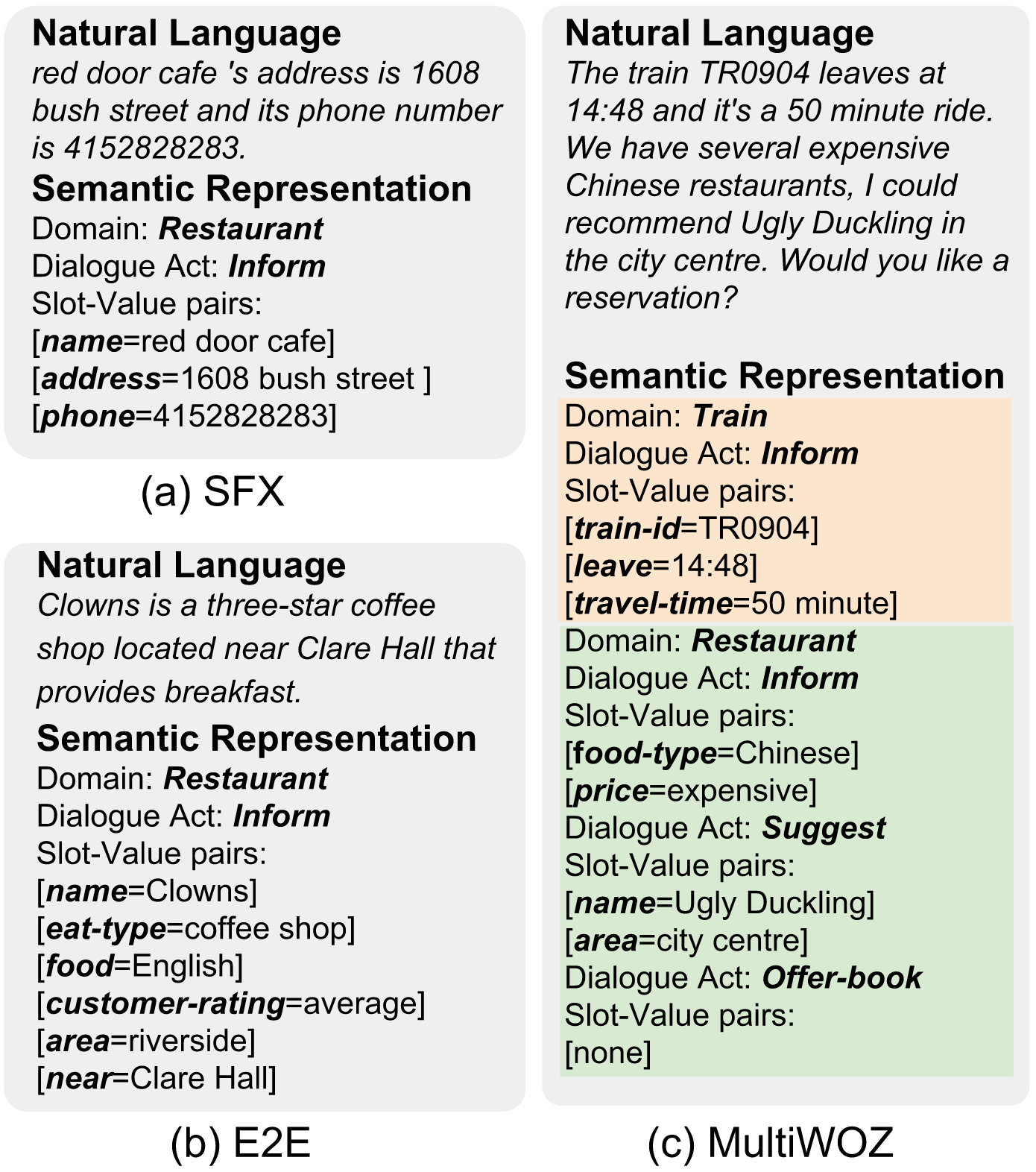}
  \caption{Examples of semantic representations in (a) SFX dataset \cite{wensclstm15}, (b) E2E dataset \cite{novikova2017e2e} and (c) MultiWOZ dataset \cite{budzianowski2018multiwoz}.}
  \label{fig:sr}
    \vspace{-0.5em}
\end{figure}

% NLG task, prior work is far way from multi-domain model (esp no knowledge sharing)
% SR examples in previous dataset and woz3
A crucial component of a Spoken Dialogue System is the Natural Language Generation (NLG) module, which generates the text that is finally presented to the user. NLG is especially challenging when building a multi-domain dialogue systems. Given a semantic representation (SR), the task for NLG is to generate natural language conveying the information encoded in the SR. Typically, an SR is composed of a set of slot-value pairs and a dialogue act consistent with an ontology. A dialogue act represents the intention of the system output and the slots provide domain-dependent information. Figure \ref{fig:sr} presents examples of SRs with their corresponding natural language representations in various datasets.
%One implicit problem of existing NLG datasets is that there is only one domain and one dialogue act per dialogue turn as shown in Figure \ref{fig:sr} (a) and (b). This limits models' ability to learn the natural language for more complex structures. For instance, the ideal system should be able to first \texttt{inform} certain information and then \texttt{request} other details from the user in the same turn. This type of example does not exist in the current dialogue datasets and, thus, cannot be generated by the current NLG models.

%\ref{fig:ontology}

% h: current position, t: top, b: bottom

% flat SR vs structured SR
%Moving to more complex dialogues spanning multiple domains exponentially increases the combinations of SR.
The input semantics has its own hierarchical structure in which there are different sets of slot-value pairs under different dialogue acts across various domains.
Modelling the semantic structure might be helpful for sharing information across domains and achieve better performance for domain adaptation.
However, prior work encodes semantic representation in a flat way such as using a binary vector \cite{wen2015stochastic,wensclstm15} or using a sequential model such as an LSTM \cite{duvsek2016sequence, tran2017natural}.
%One potential problem is that 
In that case, the structure of semantics is not fully captured by these encoding methods. This might limit models' performance especially when adapting to a new domain.

%Representing semantics by a flat feature might not be able to effectively capture the structure within the semantics.
%. The encoding that can effectively model the internal structure of SR is expected to help generator to gain better performance across multiple domains.
%Prior work encodes SR either in a binary vector \cite{wen2015stochastic,wensclstm15} or using a sequential model such as an LSTM \cite{duvsek2016sequence, tran2017natural}. One potential problem is that the structure of SR is not modeled by these encoding methods. This might limit models' performance when adapting to a new domain.

% this work
This paper investigates the possibility of leveraging the semantic structure for NLG domain adaptation in dialogue systems. We present a generation model with a tree-structured semantic encoder that models the internal structure of the semantic representation to facilitate knowledge sharing across domains. Moreover, we propose a layer-wise attention mechanism to improve the generation performance. We perform experiments on the multi-domain Wizard-of-Oz corpus (MultiWOZ) \cite{budzianowski2018multiwoz} and with human subjects. %MultiWOZ often consists of single turns with several dialogue acts with their corresponding slot-value pairs across many domains.
The results show that the proposed model outperforms previous methods on both automatic metrics and with human evaluation, suggesting that modelling the semantic structure can facilitate domain adaptation. To the best of our knowledge, this work is the first study exploiting the tree LSTM \cite{tai2015improved} to model the input semantics of NLG in spoken dialogue systems.

\section{Related Work}
% tradiational
% nn NLG
Recently, recurrent neural network-based NLG models have shown their powerful capability and flexibility compared to traditional approaches that depend on hand-crafted rules in dialogue systems.
%\cite{stent2004trainable, langkilde1998generation}.
A key development was the heuristic gate which turns off the slots that are already generated in the output sentence \cite{wen2015stochastic}. Subsequently, the semantically conditioned LSTM (SCLSTM) \cite{wensclstm15} was proposed with an extra reading gate in the LSTM cell to let the model automatically learn to control the binary representation of the semantics during generation. The sequence-to-sequence (seq2seq) model \cite{cho2014learning,sutskever2014sequence} with attention mechanism \cite{bahdanau2014neural} that has achieved huge success in machine translation has also been applied to the NLG task.
%The mapping from the semantics to natural language can be formulated as a seq2seq learning problem.
In \cite{duvsek2016sequence} the slot-value pairs in the semantics were treated as a sequence and encoded by LSTM. Based on the seq2seq model, in \cite{tran2017neural,tran2017natural} the refinement gate was introduced to modify the input words and hidden states in the decoder by considering the attention result. Different training strategies were studied in prior work. The hierarchical decoding method was proposed by considering the linguistic pattern of the generated sentence \cite{su2018natural}. The variational-based model was proposed to learn the latent variable from both natural language and semantics \cite{tseng2018variational}. \citet{lampouras2016imitation} proposed to use imitation learning to train NLG models, where the Locally Optimal Learning to Search framework was adopted to train against non-decomposable loss functions.
%The 

% adaptation NLG
Domain adaptation has been widely studied in different areas such as machine translation \cite{koehn2007experiments,foster2010discriminative}, part of speech tagging \cite{blitzer2006domain} and dialogue state tracking \cite{mrkvsic2015multi} in spoken dialogue systems. In NLG for spoken dialogue systems, the trainable sentence planner proposed in \cite{walker2002training,stent2004trainable} provides the flexibility of adapting to different domains. Subsequently, generators that can tailor user preferences \cite{walker2007individual} or learn their personality traits \cite{mairesse2008trainable,mairesse2011controlling,oraby2018controlling} were proposed. To achieve multi-domain NLG, exploiting the shared knowledge between domains is important to handle unseen semantics. A multi-step procedure to train a multi-domain NLG model was proposed in \cite{wenmultinlg16}.
%in which the model was first trained on counterfeited data and then fine turned with adaptation data.
Adversarial learning is used in \cite{tran2018adversarial} in which two critics were introduced during model adaptation.
%to learn patterns from the target domain.

\section{Model}
\begin{figure*}[htb]
  \includegraphics[width=1.0\textwidth]{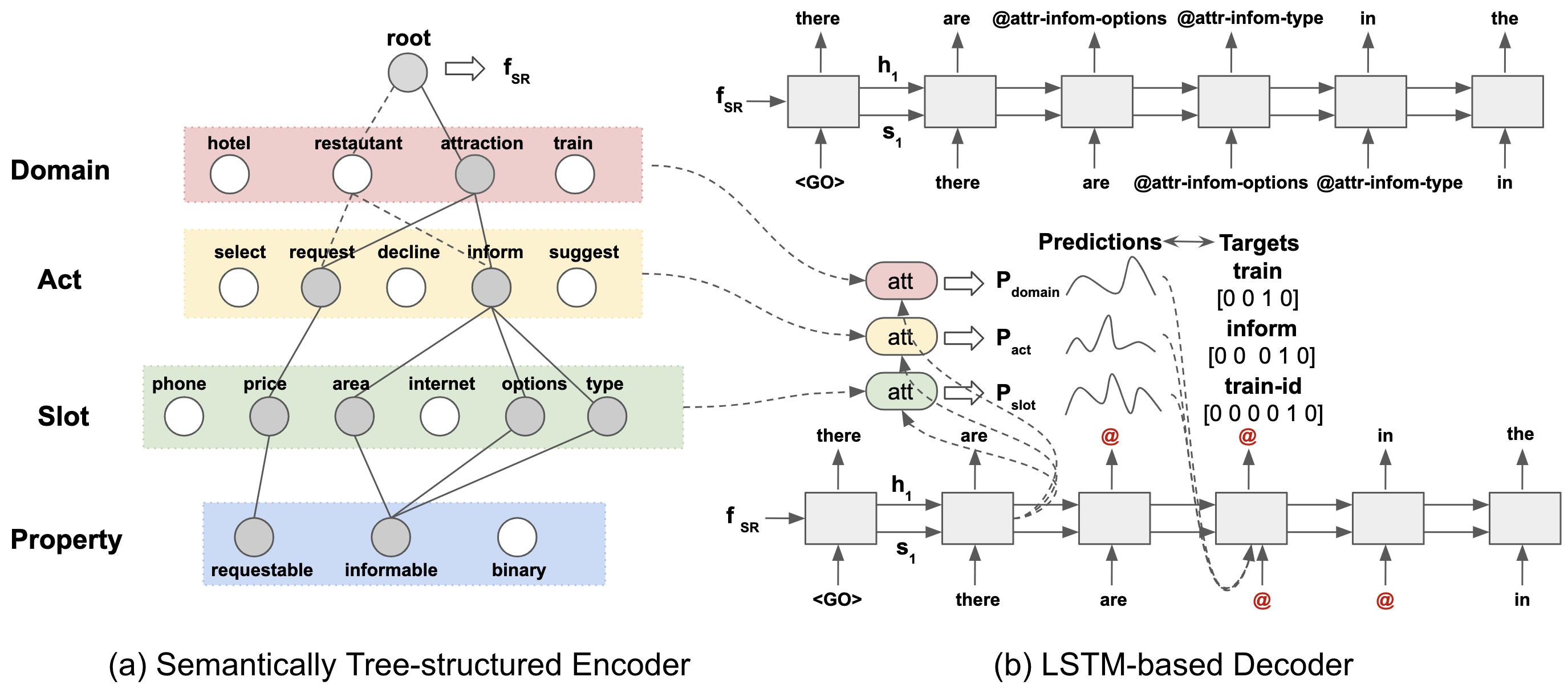}
  \caption{The overview of our generation model. The tree-structured semantic encoder (a) encodes semantic representation to obtain semantic embedding $f_{SR}$. Each node in the tree denotes a vector representation for that token. Grey node means it is activated during encoding with the corresponding token specified in the semantics. The LSTM-based decoder without layer-wise attention ((b), above) and with layer-wise attention ((b), below) takes $f_{SR}$ as an initial state to generate natural language. The example utterance here is \textit{"there are @attraction-inform-options @attraction-inform-type in the @attraction-inform-area, do you have a price range in mind?"}}
  \label{fig:model}
    \vspace{-0.5em}
\end{figure*}
% model has encoder and decoder
Our generation model is composed of two parts: (a) a tree-structured semantic encoder and (b) an LSTM decoder with additional gates. The tree-structured semantic encoder extracts a semantic embedding from the semantics in a bottom-up fashion. The obtained embedding is then fed into the decoder as a condition to generate natural language with corresponding delexicalised tokens\footnote{Each value in a natural language utterance is replaced by a delexicalised token in the format \texttt{@domain-act-slot}. For instance, the informed restaurant \textit{Golden House} will be replaced by the token \texttt{@restaurant-inform-name}. The mapping from values to delexicalised tokens is called delexicalisation. The inverse process is called lexicalisation.}. In addition, we further propose a layer-wise attention mechanism between the tree-structured semantic encoder and the decoder. The proposed attention mechanism further improves the model's ability to generate the correct information when adapting to a new domain with limited data.

\subsection{Tree-Structured Semantic Encoder}
% overview of tree-structured encoder
% types of values
% transition between layers
%comparing to 1-hot and seq encoder, what do we expect
There exists a hierarchical relationship between dialogue acts and slot-value pairs within various domains. Inspired by the tree-structured LSTM \cite{tai2015improved} that encodes natural language by capturing its syntactic properties, we propose a tree-structured semantic encoder to encode the semantic representation (SR) by exploiting its internal hierarchy.

\subsubsection{Tree Hierarchy}
Figure \ref{fig:model} (a) illustrates our tree-structured semantic encoder.
The hierarchy of the tree represents the ontology with each layer symbolizing a different level of information.
At each layer, a node denotes a possible type defined by the ontology.
Given an SR, each slot-value pair is associated with a dialogue act (DA) within a domain.
This relationship is modelled by the links between different layers in a tree as parents and children.
For instance, the node denoting slot \texttt{name} is the child of the node denoting DA \texttt{suggest} and DA \texttt{suggest} is the child of the node representing domain \texttt{restaurant}.
In addition, a slot can be \textit{requestable}, \textit{informable} or \textit{binary}.
Each of them behaves differently in natural language\footnote{For instance, the utterance with a \textit{requestable} slot \texttt{area} might be: \textit{Which part of the city you are looking for?}. The utterance with the \textit{informable} slot \texttt{area} might be: \textit{There are several restaurants in the @restaurant-inform-area}.}.
Each leaf node denotes a property that describes a slot.
As a result, given an SR there is a one-to-one mapping between SR and its corresponding tree and a path from the root to a leaf node describes a slot-value pair along with its domain, DA, slot and property of slot information.

\subsubsection{Semantic Representation Encoding}
Given a tree representing an SR, each node $j$ of the LSTM contains input, forget and output gates $i_{j}$, $f_{j}$ and $o_{j}$ respectively to obtain its hidden state and memory cell $h_{j}$ and $c_{j}$. With a set of children $C(j)$, the non-leaf node $j$ has two sources of input: (a) the token embedding $e_{j}$\footnote{All the domains, dialogue acts and slots appearing in an SR are viewed as tokens and encoded in the 1-hot vectors. The 1-hot vectors are then passed through an embedding layer to attain the token embeddings as inputs to the nodes.} and (b) children states $h_{k}$, $c_{k}$. The transition equations are as following:
% i: idx for parent, j: idx for children
\begin{align*}
    &\tilde{h}_{j} = \sum_{k \in C(j)} h_{k}, \\
    &\tilde{c}_{j} = \sum_{k \in C(j)} c_{k}, \\
    &i_{j} = \sigma(W^{(i)}_{E}e_{j}+U^{(i)}_{E}\tilde{h}_{j}+b^{(i)}_{E}), \\
    &f_{j} = \sigma(W^{(f)}_{E}e_{j}+U^{(f)}_{E}\tilde{h}_{j}+b^{(f)}_{E}), \\
    &o_{j} = \sigma(W^{(o)}_{E}e_{j}+U^{(o)}_{E}\tilde{h}_{j}+b^{(o)}_{E}), \\
    &g_{j} = \tanh (W^{(g)}_{E}e_{j}+U^{(g)}_{E}\tilde{h}_{j}+b^{(g)}_{E}), \\
    &c_{j} = i_{j} \circ g_{j} + f_{j} \circ \tilde{c}_{j}, \\
    &h_{j} = o_{j} \circ \tanh(c_{j}),
\end{align*}
where $k$ is the children index, $\tilde{h}_{j}$ and $\tilde{c}_{j}$ are the sum of children's hidden states and memory cells respectively.

The semantic embedding is obtained in a bottom-up fashion. Starting from the leaf nodes with their corresponding embeddings, the information is propagated from the property layer through the slot layer, act layer and domain layer to the root. The hidden state at the root is the final semantic embedding $f_{SR}$ for the SR and it will be used to condition the decoder during generation.

During domain adaptation, the model might have seen some semantics in source domain (denoted by dash lines in the tree encoder in Figure \ref{fig:model}) that shares a partial tree structure with the semantics in the target domain.
For instance, the SR informing about options, type and area in restaurant domain shares partial tree structure with the SR informing about the same information in attraction domain. Modelling semantic structure by the tree encoder benefits knowledge sharing across domains. 

\subsection{Decoder}
Figure \ref{fig:model} (b) presents the LSTM-based decoder with two introduced gates.
%During generation, both hidden state and semantic feature update at each time step and flow along time sequence.
%Updating the semantic feature is crucial for the model to avoid generating redundant or missing information in the SR.
The representation of the semantics, $s_{t}$, is initialised by the semantic embedding $f_{SR}$ and then updated at each time step duration generation.
Updating the semantics at each step is crucial to avoiding generating redundant or missing information in the SR.
As in standard LSTMs, the transition equations of memory cell $c_{t}$ are as following:
\begin{align*}
    &i_{t} = \sigma(W^{(i)}_{D}x_{t}+U^{(i)}_{D}h_{t-1}+b^{(i)}_{D}), \\
    &f_{t} = \sigma(W^{(f)}_{D}x_{t}+U^{(f)}_{D}h_{t-1}+b^{(f)}_{D}), \\
    &o_{t} = \sigma(W^{(o)}_{D}x_{t}+U^{(o)}_{D}h_{t-1}+b^{(o)}_{D}), \\
    &g_{t} = \tanh(W^{(g)}_{D}x_{t}+U^{(g)}_{D}h_{t-1}+b^{(g)}_{D}), \\
    &c_{t} = i_{t} \circ g_{t} + f_{t} \circ c_{t-1}.
\end{align*}
The two introduced gates, reading gate $r_t$ and writing gate $w_t$, are responsible for updating the semantic state $s_{t}$. The reading gate determines what information should be kept from the semantics at previous time step, while the writing gate decides what new information should be added into the current semantic state:
\begin{align*}
    &r_{t} = \sigma(W^{(r)}_{D}x_{t}+U^{(r)}_{D}h_{t-1}+V^{(r)}_{D}s_{t-1}+b^{(r)}_{D}), \\
    &w_{t} = \sigma(W^{(w)}_{D}x_{t}+U^{(w)}_{D}h_{t-1}+V^{(w)}_{D}s_{t-1}+b^{(w)}_{D}), \\
    &d_{t} = \tanh(W^{(d)}_{D}x_{t}+U^{(d)}_{D}h_{t-1}+V^{(d)}_{D}s_{t-1}+b^{(d)}_{D}), \\
    &s_{t} = w_{t} \circ d_{t} + r_{t} \circ s_{t-1}.
\end{align*}

The hidden state $h_{t}$ is then defined as the weighted sum of the memory cell and the semantic state with the output gate as weight:
\begin{equation*}
    h_{t} = o_{t} \circ \tanh(c_{t}) + (1-o_{t}) \circ \tanh(s_{t}).
\end{equation*}
The probability of the word label $y_{t}$ at each time step $t$ is formed by a applying a softmax classifier that takes the hidden state $h_{t}$ as input:
\begin{equation*}
    p(y_{t}|x_{<t}, f_{SR}) = \text{softmax}(W^{(s)}h_{t}).
\end{equation*}
The objective function is the standard negative log-likelihood:
\begin{equation}
\label{eq:obj}
    J(\theta) = -\sum_{t}\log p(y_{t}|x_{<t}, f_{SR}).
\end{equation}

\subsection{Layer-wise Attention Mechanism}
The semantic embedding obtained from the tree encoder contains high-level information regarding the semantic representation. However, the information in the tree is not fully leveraged during generation. Thanks to the hierarchical structure of a tree encoder with defined meaning for each layer, we can apply an attention mechanism to each layer to let the decoder concentrate on the different levels of information. We expect the decoder to leverage information regarding domain, dialogue act and slot from the hidden states in a tree to influence the generation process.

Whenever the decoder generates the token \texttt{@}\footnote{With the layer-wise attention mechanism, all values in the natural language are replaced by the same delexicalised token \texttt{@} instead of the tokens in the format \texttt{@domain-act-slot}, and the corresponding information regarding domain, dialogue act and slot will be used as signals to guide the decoder to predict the correct information.}, the semantics $s_{t}$ is used to drive an attention mechanism with hidden states in the different layers of the tree to obtain  distributions over domains $p(d_{t})$, dialogue acts $p(a_{t})$ and slots $p(s_{t})$ respectively:
\begin{align*}
    &p(d_{t}|x_{<t}, s_{t}) = \frac{\exp(\text{score}(s_{t}, h_{d}))}
                                {\sum_{d'\in D}\exp(\text{score}(s_{t}, h_{d'}))}, \\
    &p(a_{t}|x_{<t}, s_{t}) = \frac{\exp(\text{score}(s_{t}, h_{a}))}
                                {\sum_{a'\in A}\exp(\text{score}(s_{t}, h_{a'}))}, \\
    &p(s_{t}|x_{<t}, s_{t}) = \frac{\exp(\text{score}(s_{t}, h_{s}))}
                                {\sum_{s'\in S}\exp(\text{score}(s_{t}, h_{s'}))},
\end{align*}
where $h_d$, $h_a$ and $h_s$ are the hidden states of domain, dialogue act and slots in the tree encoder. $D$, $A$ and $S$ are the sets of domains, dialogue acts and slots defined in the ontology respectively. The score function used to calculate the similarity between two vectors is defined as following:
\begin{equation*}
    \text{score}(f, h) = f^{T} h.
\end{equation*}
The distributions $p(d_{t})$, $p(a_{t})$ and $p(s_{t})$ are then used to predict domain, dialogue act and slot at time step $t$ by taking the argmax operation to form the delexicalised tokens \texttt{@domain-act-slot} back into the generated sentence.

In order to avoid generating redundant or missing information in a given SR, the three predicted distributions are fed into next time step to augment the original input word
\footnote{Only at the next time step of generating delexicalised token \texttt{@} the input is the concatenation of the word vector $x_{t}$ and three predicted distributions. In any other time steps, the input is the word vector padded with zeros.}
%\footnote{Only at the next time step of predicting the domain, act and slot the input is augmented. In any other time step during generation, the input is simply the predicted word from the previous time step.}
to condition the model on what information has already been generated.
%\footnote{We conducted the experiments without augmented inputs from the predictions regarding domain, act and slot. The results are worst.}.

During training, the error signals between predicted distributions and the true labels for domain, dialogue act and slot are added to the objective function. The objective function for the generation model with layer-wise attention mechanism is defined as following:
\begin{align*}
    J_{att}(\theta) = J(\theta)-\sum_{t'} \left(\log p(d_{t'}|x_{<t'}, s_{t'}) \right. \\
   \left. +\log p(a_{t'}|x_{<t'}, s_{t'})+\log p(s_{t'}|x_{<t'}, s_{t'}) \right),
\end{align*}
where $J(\theta)$ is the original objective function in equation \ref{eq:obj} and $t'$ is the index for the time step where each token \texttt{@} is generated.

% 5 pages above
\section{Experimental Results}
\subsection{Dataset} %0.5 page
% woz3, how to collect, compare with other dataset how difficult it is
We perform our experiments with the Multi-Domain Wizard-of-Oz (MultiWOZ) dataset \cite{budzianowski2018multiwoz} that is a rich dialogue dataset spanning over 7 domains. There are $10438$ dialogues and over $115k$ turns in total. The dataset contains a high level of complexity and naturalness which is suitable for developing multi-domain NLG models. There are multiple utterances in a single turn with an average of 18 words, 1.6 dialogue acts and 2.9 slots per turn.
Some turns provide information for more than 1 domain. Comparing with previous NLG datasets which contain only 1 utterance in a turn with 1 dialogue act within 1 domain, the MultiWOZ dataset provides significantly more complexity and makes NLG more challenging. The number of examples, distinct semantic representation (SR) and numbers of dialogue acts and slots are reported in Table \ref{tab:statistics}. The data split for train, dev and test is 3:1:1. The details of the ontology is presented in Figure \ref{fig:ontology}.

%We observe that there are more than 12k dialogue turns with only \texttt{good-bye}, \texttt{welcome} or \texttt{greeting} as dialogue act. As these utterances are not difficult to generate correctly, they usually lead to an artificially good result on the automatic evaluation such as BLEU that compares the similarity between two utterances. Therefore, we remove these turns and the turns with potentially wrong labels from the original dataset in our experiments. The number of examples, distinct semantic representation (SR) and numbers of dialogue acts and slots are reported in Table \ref{tab:statistics}. The data split for train, dev and test is 3:1:1. The details of the ontology is presented in Figure \ref{fig:ontology}.

\begin{table}[tb]
\centering
\caption{The data statistics for each domain.}
\label{tab:statistics}
\resizebox{\columnwidth}{!}{%
\begin{tabu}{l|ccccc}
\tabucline [1pt]{1-7}
\textbf{Domain} & \textbf{Restaurant} & \textbf{Hotel} & \textbf{Attraction} & \textbf{Train} & \textbf{Taxi} \\ \tabucline [1pt]{1-7}
Examples & 8.5k & 6.6k & 6.4k & 11k & 3.4k \\
Distinct SR & 346 & 378 & 314 & 338 & 47 \\
Dialogue acts & 8 & 8 & 8 & 5 & 2 \\
Slots & 11 & 14 & 13 & 11 & 6 \\
\tabucline [1pt]{1-7}
\end{tabu}%
}
\end{table}

\begin{figure*}[htb]
  \includegraphics[width=\linewidth]{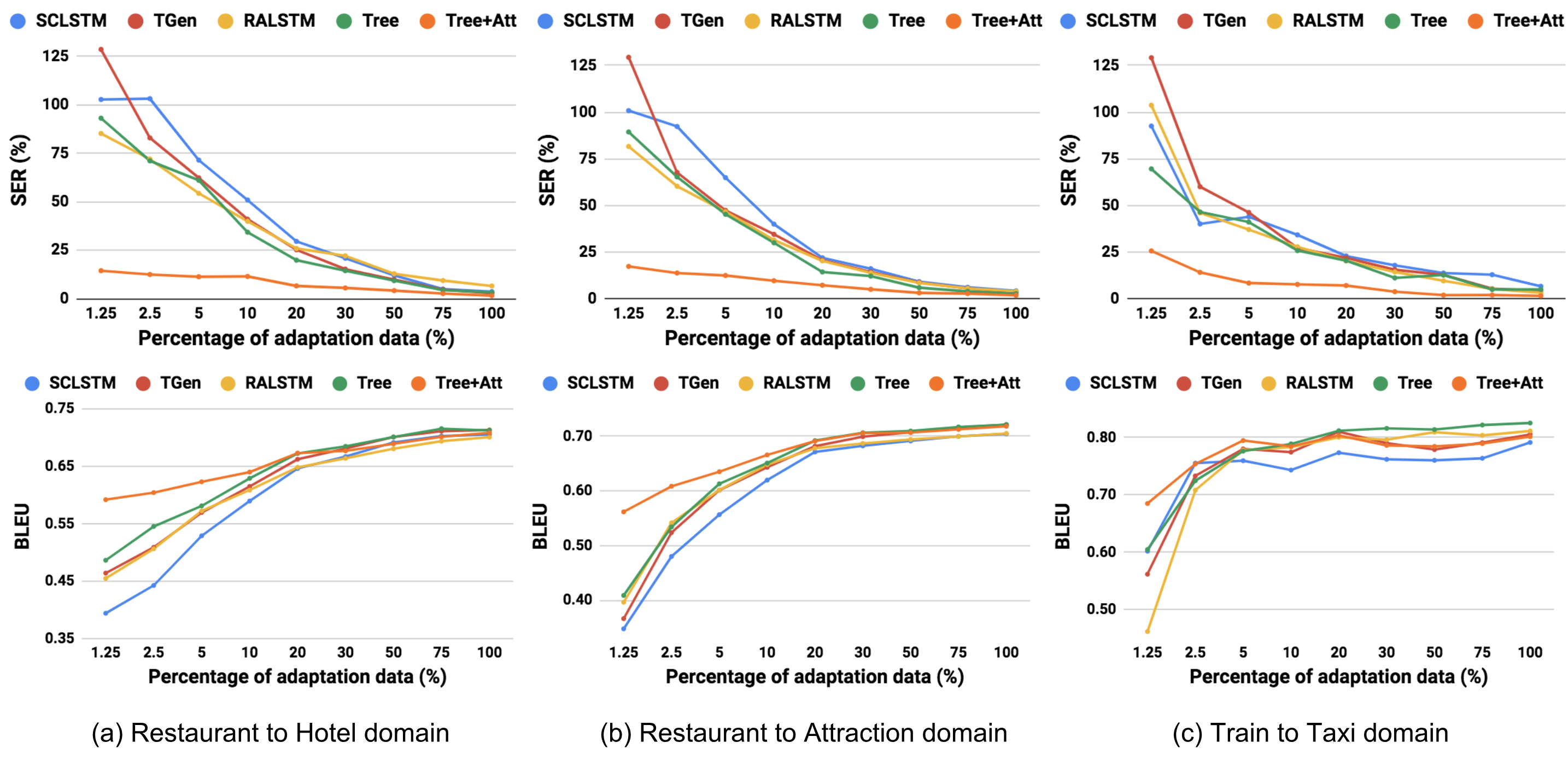}
  \caption{Domain adaptation experiments in three different settings. (a) adapting to hotel from restaurant domain. (b) adapting to attraction from restaurant domain. (c) adapting to taxi from train domain.}
  \label{fig:res_adapt}
    \vspace{-0.5em}
\end{figure*}

\subsection{Experimental Setup} %0.5 page
% hyper-parameters, how to evaluate, e.g., generate 10 different samples and avg. 5 seeds. please not the match for slot
The generators are implemented using the Pytorch library \cite{paszke2017automatic}%. Our code is public\footnote{\url{https://github.com/andy194673/TreeEncoder-NLG-Dialogue}}.
\footnote{The code will be released soon on github}.
The number of hidden units in the LSTMs is 100 with 1 hidden layer. The dropout rate is 0.25 and the Adam optimizer is used. The learning rate is 0.0025 for the models trained from scratch, and 0.001 for the models adapted from one domain to another in adaptation experiments. Beam search is used during decoding with a beam size 10. For automatic metrics, the BLEU scores and the slot error rate (SER) used in \cite{wensclstm15} are reported. The SER is used to evaluate how accurate a generated sentence is in terms of conveying the desired information in the given semantic representation (SR). The SER is defined as: $(p+q)/N$, where $p, q$ are the numbers of missing and redundant slots in a generated sentence, and $N$ is the number of total slots that a generated sentence should contain. The results are averaged over 10 samples and 5 random initialised seeds. As explained above each delexicalised slot token in an utterance is in the format of \texttt{@domain-act-slot}. When calculating the SER, the predicted slot token is correct only if its domain, dialogue act and slot information are all correct. For example, if there is a desired slot \texttt{area} under dialogue act \texttt{inform} within \texttt{restaurant} domain in SR, the model needs to generate the token \texttt{@restaurant-inform-area}. 

The tree-structured semantic encoder (Tree) and the variant with attention (Tree+Att) are compared against three baselines: (1) the semantically-conditioned LSTM (SCLSTM) that has an extra gate to update the binary vector of the semantic representation \cite{wensclstm15}; (2) TGen that is a seq2seq model with attention mechanism mapping SR into a word sequence \cite{duvsek2016sequence}; (3) a refinement adjustment LSTM (RALSTM) that is an improved seq2seq model with a refinement gate and an adjustment gate in the decoder \cite{tran2017natural}.

As the decoding method is slightly different between our model Tree+Att and baseline models\footnote{Tree+Att only generates token \texttt{@} and reply on attention results to form the complete slot token while baseline models directly generate slot tokens.}, in order to guarantee the optimised baseline systems, we also trained baseline models in the same decoding way as Tree+Att to only predict \texttt{@} with three additional classifiers for domain, act and slot prediction. However, baseline models obtains better performance by the original decoding method so we keep that in the following experiments. All the models are optimized by selecting the best one based on the validation set result.

\subsection{Automatic Evaluation}
In order to examine the models' ability to share knowledge between domains, we performed experiments in three domain adaptation scenarios: (a) adapting to hotel from restaurant domain; (b) adapting to attraction from restaurant domain and (c) adapting to taxi from train domain. The adaptation models were fine-tuned with adaptation data based on the models trained on source domain\footnote{All the multi-domain turns are removed in case the model have seen any examples related to target domain before adaptation.}. The SER results are presented in the first row of Figure \ref{fig:res_adapt}. Generally, our model without attention (Tree) performs similarly with RALSTM but better than TGen and SCLSTM. With the layer-wise attention mechanism, our model (Tree+Att) improves significantly and performs better than baselines at all different levels of adaptation data amount. Especially when the adaptation data used is only 1.25\%, the SER is reduced from above 75\% to around 25\%. We found that this is because baseline models tend to predict the slots with the wrong dialogue act or in the wrong domain as the limited adaptation data
%\footnote{Only 50 utterances were used in scenario (a) (b) and 20 utterances were used in scenario (c) during adaption.}
makes it difficult to learn the sentence pattern in the target domain. However, with the layer-wise attention mechanism, our model is able to pay attention on the information at different levels in the tree to make the correct predictions. (See more details in section 5 with error analysis and visualisation of attention distributions.)
%Furthermore, even though SCLSTM and TGen perform well in the single-domain comparison (section \ref{sec:single}), they do not provide good performance in the adaptation experiments. This indicates that binary encoding or sequential encoding for semantic representations do not allow knowledge to be transferred from one domain to another efficiently.
%However, our models shows a powerful capability of sharing knowledge between domains during adaptation and remains a accepted-well performance in the general experiment.
A similar trend can be observed in the BLEU results in the second row of Figure \ref{fig:res_adapt}.

% use tabu
%\tabucline [1pt]{1-7}
\begin{table}[tb]
\centering
\caption{Human evaluation for utterance quality in three adaptation settings: Restaurant (Rest.) to Hotel domain; Restaurant to Attraction (Attr.) domain and Train to Taxi domain. Informativeness (Info.) and Naturalness (Nat.) are reported (rating out of 5). }
\label{tab:human-info}
\resizebox{\columnwidth}{!}{%
%\begin{tabular}{lcc||cc||cc}
\begin{tabu}{l|cc||cc||cc}
\tabucline [1pt]{1-7}
\multirow{2}{*}{\textbf{Model}} & \multicolumn{2}{c||}{\textbf{Rest. to Hotel}} & \multicolumn{2}{c||}{\textbf{Rest. to Attr.}} & \multicolumn{2}{c}{\textbf{Train to Taxi}} \\
                                & \textbf{Info.}        & \textbf{Nat.}        & \textbf{Info.}        & \textbf{Nat.}        & \textbf{Info.}       & \textbf{Nat.}       \\
\tabucline [1pt]{1-7}
SCLSTM                          & 2.96                  & 3.85                 & 2.81                  & 3.69                 & 3.05                 & \textbf{4.26}       \\
TGen                            & 2.87                  & 3.33                 & 3.00                  & 3.23                 & 3.42                 & 3.90                \\
RALSTM                          & 2.79                  & 3.48                 & 2.91                  & 3.40                 & 3.48                 & 3.15                \\  \hline
Tree                            & 3.08                  & 3.54                 & 3.38                  & 3.41                 & 3.81                 & 3.81                \\  
Tree+Att                        & \textbf{4.04}         & \textbf{4.10}        & \textbf{4.30}         & \textbf{3.92}        & \textbf{4.29}        & 3.78    \\
\tabucline [1pt]{1-7}
%\end{tabular}%
\end{tabu}%
}
\vspace{-0.5em}
\end{table}

\subsection{Human Evaluation} %0.5 page
Because automatic evaluation such as BLEU may not consistently agree with human perception \cite{stent2005evaluating}, we performed human testing via the Amazon Mechanical Turk service. We showed MTurk workers the generated sentences in adaptation experiments with adaptation data from $1.25\%$ to $10\%$ as we focus on the models' performance with limited adaptation data. Five models were compared together by showing, for each model, the 2 sentences with the highest probabilities out of the 10 generated sentences by beam search. 
The workers were asked to score each sentence from 1 (bad) to 5 (good) in terms of its informativeness and naturalness. The \textit{informativeness} is defined as the degree to which the generated sentence contains all the information specified in the given semantic representation (SR) without conveying extra information and the \textit{naturalness} is defined as whether the sentence is natural like human language.
%In addition, workers were asked to choose one model they preferred the most and one they preferred the least.
\citet{ipeirotis2010quality} pointed out that malicious workers might take advantage of the difficulty of verifying the results and therefore submit answers with low quality. In order to filter out submissions with bad quality, we also asked them to score the ground truth sentence and an artificial sentence containing irrelevant information to the SR. If the worker gave ground truth sentence a low score ($< 3$) or gave the artificial sentence a high score ($> 3$) in terms of informativeness, the submission was discarded.

% \usepackage{graphicx}
\begin{table*}[htb]
\centering
\caption{Example semantic representations (SR) with ground truth sentences in two adaptation settings with 1.25\% adaptation data and the top-1 sentences generated by each model. Both are adapted from restaurant domain. The slot-value pairs are in bold. Each generated sentence is followed by a brief description to explain if the sentence correctly conveys the information in the SR.}
\label{tab:ex}
\resizebox{\textwidth}{!}{%
\begin{tabu}{l|c|c}
\tabucline [1pt]{1-7}
Domain                                                            & \textbf{Attraction}                                                                                                                                                                                            & \textbf{Hotel}                                                                                                                                                                                     \\ \hline
\begin{tabular}[c]{@{}l@{}}Semantic\\ Representation\end{tabular} & \begin{tabular}[c]{@{}c@{}}Dialogue Act: \texttt{Inform}\\ Slot-Value pairs: {[}Area: \textbf{west}{]}\\ {[}Options: \textbf{five}{]} {[}Type: \textbf{colleges}{]}\\ Dialogue Act: \texttt{Request}\\ Slot-Value pairs: {[}Price=?{]}\end{tabular} & \begin{tabular}[c]{@{}c@{}}Dialogue Act: \texttt{Inform}\\ Slot-Value pairs: {[}Options=\textbf{two}{]}\\ Dialogue Act: \texttt{Select}\\ Slot-Value pairs: \\ {[}type1=\textbf{guesthouse}{]} {[}type2=\textbf{hotel}{]}\end{tabular} \\ \hline
Ground Truth                                                      & \begin{tabular}[c]{@{}c@{}}\textit{there are \textbf{five} \textbf{colleges} in the \textbf{west}.}\\ \textit{do you mind paying an entrance fee ?}\end{tabular}                                                                                  & \begin{tabular}[c]{@{}c@{}}\textit{i have \textbf{two}, would you} \\ \textit{prefer a \textbf{guesthouse} or \textbf{hotel} ?}\end{tabular}                                                                                           \\ \hline \hline
SCLSTM                                                            & \begin{tabular}[c]{@{}c@{}}\textit{what type of place are you looking for ?}\\ \textsf{(miss 3 slots \& request wrong)}\end{tabular}                                                                                    & \begin{tabular}[c]{@{}c@{}}\textit{what area would you} \\ \textit{like to stay in ?} \textsf{(miss 3 slots)}\end{tabular}                                                                                           \\ \hline
TGen                                                              & \begin{tabular}[c]{@{}c@{}}\textit{there are located in the . do you have a price}\\ \textit{range in mind ?} \textsf{(miss 3 slots)}\end{tabular}                                                              & \begin{tabular}[c]{@{}c@{}}\textit{i have found options. would you} \\ \textit{prefer or ?} \textsf{(miss 3 slots)}\end{tabular}                              \\ \hline
RALSTM                                                            & \begin{tabular}[c]{@{}c@{}}\textit{we have \textbf{five} \textbf{colleges} in the \textbf{west} area . do you} \\ \textit{have an attraction type in mind ?} \textsf{(request wrong)}\end{tabular}                                                           & \begin{tabular}[c]{@{}c@{}}\textit{i have \textbf{two} options.} \\ \textit{do you have a preference ?} \textsf{(miss 2 slots)}\end{tabular}                                                                                  \\ \hline
Tree                                                              & \begin{tabular}[c]{@{}c@{}}\textit{there are \textbf{five} \textbf{colleges} in the \textbf{west} . do you have an}\\ \textit{area of town you would prefer ?} \textsf{(request wrong)}\end{tabular}                                                         & \begin{tabular}[c]{@{}c@{}}\textit{i have found \textbf{two} options for you.} \\ \textit{do you have a preference ?} \textsf{(miss 2 slots)}\end{tabular}                                                                    \\ \hline
Tree+Att                                                          & \begin{tabular}[c]{@{}c@{}}\textit{there are \textbf{five colleges} in the \textbf{west} .} \\ \textit{do you have a price range in mind ?} \textsf{(correct)}\end{tabular}                                                                         & \begin{tabular}[c]{@{}c@{}}\textit{i have \textbf{two} options for you. would} \\ \textit{you prefer \textbf{guesthouse} or \textbf{hotel} ?} \textsf{(correct)}\end{tabular}                                                \\       
\tabucline [1pt]{1-7}
\end{tabu}%
}
\end{table*}

\begin{table}[tb]
\centering
\caption{Error analysis - number of examples in the testing set and the number of wrong generated utterances (at least 1 missing or redundant slot) by each model in different adaptation data scenarios. The testing example is defined as seen if its semantics appears in the training set.}
%If the semantics of a testing example appears in the training set, this example is marked as seen. Otherwise, the example is marked as unseen.}
\label{tab:analysis}
%\resizebox{\textwidth}{!}{%
\resizebox{\columnwidth}{!}{%
\begin{tabu}{l|cc||cc||cc||cc}
\tabucline [1pt]{1-9}
\textbf{Percentage}                                                                       & \multicolumn{2}{c||}{\textbf{1.25\%}} & \multicolumn{2}{c||}{\textbf{5\%}} & \multicolumn{2}{c||}{\textbf{10\%}} & \multicolumn{2}{c}{\textbf{50\%}} \\
\tabucline [1pt]{1-9}
\multirow{2}{*}{\begin{tabular}[c]{@{}l@{}}\textbf{Testing}\\ \textbf{examples}\end{tabular}} & \textbf{seen}         & \textbf{unseen}        & \textbf{seen}       & \textbf{unseen}       & \textbf{seen}        & \textbf{unseen}       & \textbf{seen}        & \textbf{unseen}       \\
                                                                                      & 439          & 902           & 858        & 483          & 1069        & 272          & 1330        & 11           \\
\tabucline [1pt]{1-9}
\textbf{SCLSTM}                                                                                & 248          & 729           & 307        & 412          & 302         & 190          & 111         & 5            \\
\textbf{TGen}                                                                                  & 309          & 741           & 176        & 353          & 178         & 168          & 102         & 6            \\
\textbf{Tree+Att}                                                                              & 10           & 134           & 31         & 103          & 60          & 55           & 76          & 3          
\\
\tabucline [1pt]{1-9}
\end{tabu}%
}
\end{table}

The results pertaining to informativeness and naturalness are reported in Table \ref{tab:human-info} in three adaptation settings: Restaurant (Rest.) to Hotel domain; Restaurant to Attraction (Attr.) domain and Train to Taxi domain. For informativeness, our models (both Tree+Att \& Tree) outperform all baseline models in the different settings. This result is consistent with the slot error rate of the automatic evaluation reported in Figure \ref{fig:res_adapt} and indicates that the tree-structured semantic encoder does help the model to produce utterances with the correct information. For naturalness, Tree+Att performs the best in two settings, while SCLSTM performs better when adapting to taxi domain. This might be because SCLSTM is good at generating utterances with simple patterns and the taxi domain is relatively easy due to its low number of combinations of SR\footnote{There are only 2 dialogue acts and 6 slots in taxi domain.}. When adapting to more complex domains such as hotel or attraction, our models provide both informative and natural utterances. Table \ref{tab:ex} presents example semantic representations with corresponding ground truth sentence and the top-1 utterance generated by each model.

\begin{comment}
In addition, we asked workers to vote for which model they preferred the most and one they preferred the least. This forms the preference distributions over models in each adaptation setting. The result is reported in Table \ref{tab:human-prefer}. Again, our models were preferred by the judges. The workers preferred Tree+Att the most and dislike Tree the least.
As many as half of the workers preferred Tree+Att the most when adapting from restaurant to attraction domain. Table \ref{tab:ex} presents example semantic representations with corresponding ground truth sentence and the top-1 utterance generated by each model.
\end{comment}

\section{Error Analysis and Observation}
In order to investigate what type of testing data our model performs better on, we divide all test set into two subsets - \emph{seen} and \emph{unseen}. If the semantics of a testing example appear in the training set, the example is defined as \emph{seen}. Otherwise, the example is marked as \emph{unseen}. Table \ref{tab:analysis} reports the number of seen and unseen examples and the number of wrong utterances (at least 1 missing or redundant slot) generated by each model with different amount of adaptation data when adapting from restaurant to hotel domain. With more adaptation data, more SRs of testing examples appear in the training set. We observe that our model obtains better generalisation ability for unseen SRs. For instance,  with 1.25\% adaptation data, Tree+Att generates 134 wrong utterances out of 902 unseen semantics (14.8\%). However, the baseline models such as SCLSTM produces 729 wrong sentences out of 902 semantics (80.5\%). We hypothesize that our model is more capable of learning sentence patterns from source domain and generate correct content for domain adaptation. For example, when adapting from restaurant to hotel domain (see Table \ref{tab:ex} - Hotel column), Tree+Att correctly learns to generalize from the training sentence: "\textit{i have two options for you, would you prefer American or Chinese}" in restaurant domain. However, SCLSTM fails to produce a similar sentence pattern.

\begin{figure}[tb]
  \includegraphics[width=\linewidth]{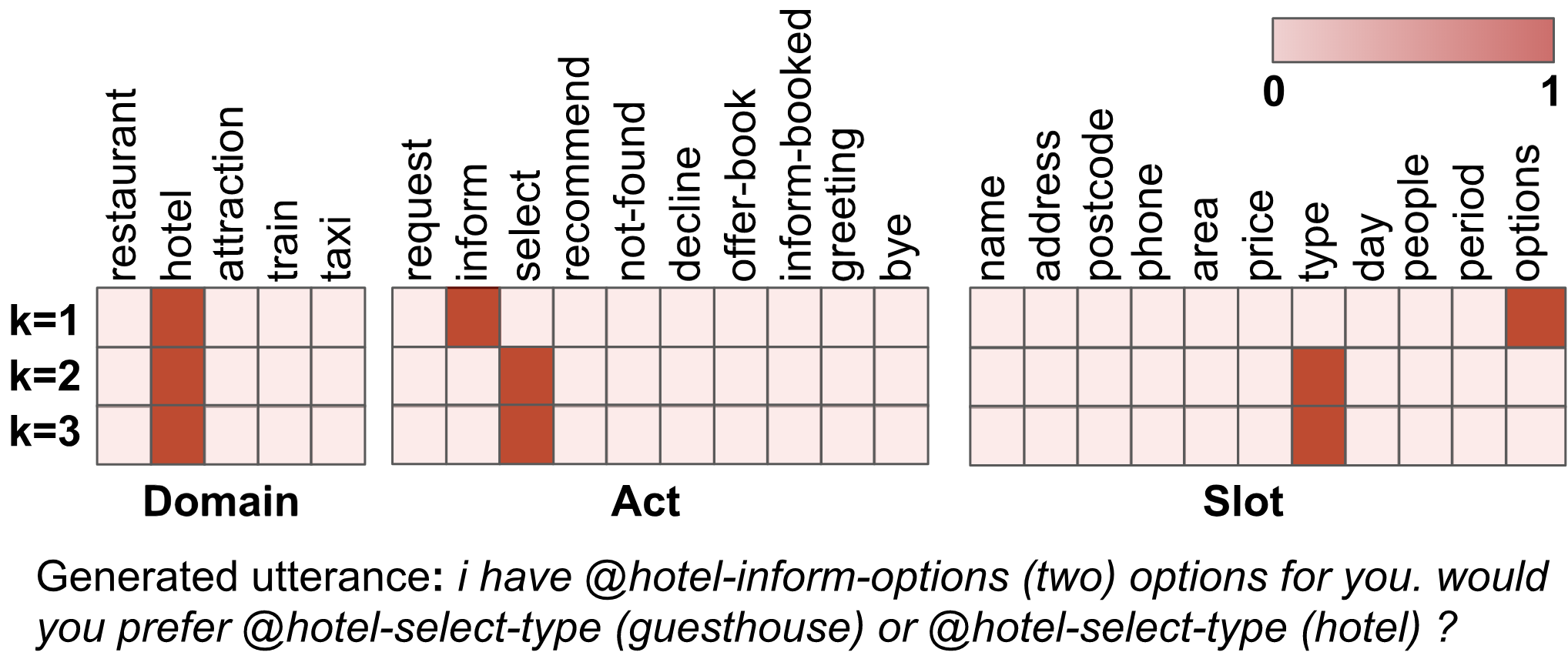}
  \caption{The visualisation of the layer-wise attention distributions over domains, acts and slots at each time step $k$ when slot token is generated and the generated utterance with lexicalised values in the parentheses. The color shades signify the attention weight.}
  \label{fig:attn}
  \vspace{-0.5em}
\end{figure}

Figure \ref{fig:attn} shows the example of visualisation of layer-wise attention distributions over domains, acts and slots generated by the Tree+Att model. The model is confident of generating the correct slot tokens with the distinct peaks indicated by the dark red color in the attention distributions even though the adaptation data used is simply 1.25\%.

\section{Conclusion and Future Work} %0.5 page
This paper investigates the possibility of leveraging internal structure of input semantics for NLG domain adaptation in dialogue systems.
The proposed tree-structured semantic encoder is able to capture the structure of semantic representations and facilitate knowledge sharing across domains. In addition, we have proposed a layer-wise attention mechanism between the tree-structured semantic encoder and the decoder to enhance the performance. Our proposed model was evaluated on the complex multi-domain MultiWOZ dataset.
%and both automatic evaluation and human assessment were conducted.
The automatic evaluation results show that our model is more efficient in terms of adaptation data usage and outperforms previous methods by reducing the slot error rate up to 50\% when the adaptation data is limited. What is more, human judges rate our model more highly than previous methods. Future work will explore a tree encoder exploiting both semantic representation and context information in end-to-end dialogue systems.

\section*{Acknowledgments}
Bo-Hsiang Tseng is supported by Cambridge Trust and the Ministry of Education, Taiwan. This work was partly funded by an Alexander von Humboldt Sofja Kovalevskaja grant.

%\noindent \textbf{Preparing References:} \\
%Include your own bib file like this:
%\verb|\bibliographystyle{acl_natbib}|
%\verb|\bibliography{acl2019}| 

%where \verb|acl2019| corresponds to a acl2019.bib file.
\bibliography{acl2019}
\bibliographystyle{acl_natbib}

% using bbl
%\nocite{*} % to test all bib entrys
%\bibliographystyle{unsrt}
%\bibliography{acl2019} % file mwe.bib

\end{document}